\documentclass[sigconf]{acmart}
\AtBeginDocument{%
  }

\copyrightyear{2023}
\acmYear{2023}
\acmConference[AEJMC]{Association for Education in Journalism and Mass Communication Conference}{August 07--10, 2023}{Washington, DC}
\acmBooktitle{the Association for Education in Journalism and Mass Communication Conference,
 August 07--10, 2023}
\usepackage{titlesec}
\usepackage{graphicx}
\usepackage{longtable}
\usepackage{geometry} % optional, for margin control
\usepackage{multirow}
\usepackage{array}
\usepackage{ragged2e} % if you're using \raggedright

\begin{document}

\title{Semantic-based Unsupervised Framing Analysis (SUFA): A Novel Approach for Computational Framing Analysis}


% \affiliation{%

\author{Mohammad Ali}
\affiliation{%
  \institution{College of Information}
  \institution{University of Maryland}
  \city{College Park}
  \country{USA}}
\email{mali24@umd.edu}
\email{aliusacomm@gmail.com}

% \affiliation{%

\author{Naeemul Hassan}
\affiliation{%
  \institution{Philip Merrill College of Journalism}
  \institution{College of Information}
  \institution{University of Maryland}
  \city{College Park}
  \country{USA}}
\email{nhassan@umd.edu}

% \affiliation{%
% \affiliation{%

% \affiliation{%

% \affiliation{%

% \affiliation{%

% \affiliation{%
%   \city{make}

\renewcommand{\shortauthors}{Ali and Hassan}

\begin{abstract}
 This research presents a novel approach to computational framing analysis, called \textit{Semantic Relations-based Unsupervised Framing Analysis} (SUFA). SUFA leverages semantic relations and dependency parsing algorithms to identify and assess entity-centric emphasis frames in news media reports. This innovative method is derived from two studies—qualitative and computational—using a dataset related to gun violence, demonstrating its potential for analyzing entity-centric emphasis frames. This article discusses SUFA’s strengths, limitations, and application procedures. Overall, the SUFA approach offers a significant methodological advancement in computational framing analysis, with its broad applicability across both the social sciences and computational domains.
 \end{abstract}




\keywords{Computational framing analysis, semantic relations, dependency parsing, natural language processing, communication method, computational method}



\maketitle

\section{Introduction}

Frames are predominantly explored using qualitative methods ~\citep[e.g.,][]{morin2016framing} and quantitative methods ~\citep[e.g.,][]{mckeever2022gun} through manual labor and analysis of small datasets. The recent proliferation of online news reports and social media posts has resulted in the generation of a vast amount of digital data that is difficult to analyze manually. To overcome this challenge, scholars have started using various computational methods, broadly divided into two parts: supervised and unsupervised ~\cite{ali2022survey}. The supervised methods require pre-determined labels and substantial human labor, while the unsupervised methods that this current research focuses on need little human effort and are applicable across domains. 

Existing unsupervised methods (e.g., topic modeling) in framing analysis mainly rely on the frequency and co-occurrence of words, leading to the exploration of topics instead of deeper framing insights ~\cite{nicholls2021computational, ali2022survey, entman1993framing}. An improved unsupervised computational solution to this longstanding communication challenge is becoming essential in this era of big data. Scholars ~\citep[e.g.,][]{ali2022survey} advocate for methods to capture semantic relationships between words, moving beyond the traditional bag-of-words approach to enhance the methodological framework. In response to these calls, this article examines semantic relationships between words, presenting a novel unsupervised approach for computational framing analysis based on dependency parsing, a natural language processing (NLP) technique largely overlooked in framing analysis. 

This mixed-method article involves two studies. Study 1 employs a qualitative textual analysis to inductively examine a sample of news reports published by four major U.S. news media outlets on the 2022 Uvalde school mass shooting in Texas, as a case study. While the political impasse and public debate continue over gun violence, it is important to understand how news media outlets frame the issue, as media framing determines how people “choose to act upon [the problem]” ~\cite[p. 54]{entman1993framing}. Study 1 examines how individual words, such as adjectives and adverbs, convey different meanings related to the shooter, victims, and the shooting event. This helps us understand how these words and their semantic relationships work together to construct frames. Study 2 employs the computational technique of dependency parsing to analyze the same dataset. Specifically, we investigate dependency parsing, along with word embedding, k-means clustering, and manual input, establishing this method as a viable approach for capturing semantic relationships and analyzing the entity-centric emphasis frames. 

Integrating qualitative and quantitative approaches in this project provides complementary strengths essential for developing the methodological approach. The qualitative analysis in Study 1 offers interpretive depth by first manually uncovering whether and how specific words and their semantic relationships contribute to frame construction in natural language. This inductive insight helps ground the methodological design. Through quantitative computational techniques, Study 2 validates and extends this insight from Study 1 by systematically extracting these patterns computationally. Together, the two studies demonstrate that semantic structures, captured through dependency parsing, can reliably identify emphasis frames, laying the foundation for a scalable, unsupervised computational framing analysis model.

The outcomes of both studies are discussed. Importantly, this mixed-method project solidified and proposed the semantic relations-based approach for framing analysis, named “Semantic Relations-based Unsupervised Framing Analysis.” The step-by-step procedure for applying this approach, along with its strengths, limitations, and future research directions, is also discussed.

% \begin{figure}[!ht]
    % \centering
    % \includegraphics[width=0.47\textwidth]{NYT-G_news_headlines2}
    % \caption{Framing devices deployed in the headlines of two news reports published by \textit{The New York Times} and \textit{The Guardian} on the 2022 Buffalo mass shooting.}
    % \label{fig:method}
% \end{figure}

\section{Literature Review}
\subsection{Framing}
Scholars have not reached a consensus on a unified definition of framing ~\cite{goffman1974frame, hertog2001multiperspectival}. However, one of the most widely cited definitions in framing studies comes from ~\citet{entman1993framing}, who posits:

\begin{quote}
To frame is to \textit{select some aspects of a perceived reality and make them more salient in a communicating text, in such a way as to promote a particular problem definition, causal interpretation, moral evaluation, and/or treatment recommendation} for the item described. (p. 52)
\end{quote}

In the news media context, a frame is “a central organizing idea” ~\cite{tankard1991media}, and it “denotes how journalists, their sources, and audiences work within conditions that shape the messages they construct as well as the ways they understand and interpret these messages” ~\cite[p. xxiv]{d2018doing}. Going beyond the idea of a simple topic, news framing is “like moving a telescope into position” ~\cite[p. 125]{fairhurst2005reframing}, where selected aspects are coherently organized in a way that makes an argument, promoting a particular interpretation, evaluation, and solution ~\cite{fairhurst2005reframing}. Importantly, \textit{a frame “operates by selecting and highlighting some features of reality while omitting others”} ~\cite[p. 53]{entman1993framing}. Echoing with this, ~\citet{fairhurst1996art} notes that a frame is \textit{“to choose one particular meaning (or set of meanings) over another”} (p. 3). 

\subsection{Emphasis vs. Equivalency Framing}

The concept of framing revolves around two broad competing aspects: emphasis framing and equivalency framing. Equivalency framing involves presenting two or more alternatives with logically equivalent phrases (e.g., loss versus gain) ~\citet{kahneman1984choices,levin1998all}. In contrast, emphasis framing refers to the act of repeatedly highlighting or associating certain pieces of information about an issue or topic, while omitting other relevant aspects ~\citet{D'Angelo2017}. This article focuses on analyzing emphasis framing with the newly proposed computational approach.

\subsection{Words in Constructing Frames}

Scholars have long identified words and phrases that construct frames. Prior studies revealed that using certain words helps identify frames ~\cite{entman1993framing, fairhurst1996art, gamson1989media, hertog2001multiperspectival}. For example, “the use of \textit{baby} versus \textit{fetus} signals a very different approach to the topic of abortion” ~\cite[p. 150]{hertog2001multiperspectival}. Prior framing studies looked at various parts of speeches, including verbs, adverbs, and adjectives, which enhances researchers’ ability to identify frame boundaries and relationships ~\cite{hertog2001multiperspectival}. The frequent use of verbs such as “falsifying,” “forging,” and “manipulating” was found to have been utilized in news reports to frame scientists ~\cite{boesman2018driving}. News reporters also use various verbs of attribution (e.g., accused, charged, blamed) to create worth for one person while devaluing another ~\cite{dickerson2001framing}. 

\subsection{Conceptualization and Operationalization of the Framing Component }

Prior studies provide evidence for using words in constructing frames ~\cite{hertog2001multiperspectival, miller1995wordnet}. When a particular word is selected or coded as part of a frame, this word directly or indirectly operates in relation to other words to express the intended framing meaning. In other words, framing meanings are often produced not by isolated words but through their associative use with surrounding words, particularly when an entity is modified by adjectives, adverbs, or verbs. 

For example, ~\citet{bantimaroudis2001covering} reported how Somali leaders were framed by U.S. news media through the repeated use of the term “warlords” in contrast to their opposition, the United Nations forces. They interpreted the frame by exposing how extensively the word “warlords” was used in the news media coverage. This current research argues that the word “warlords” alone does not sufficiently convey a practical meaning for understanding the frame about Somali leaders. Instead, we better understand the intended frame when the word “warlords” is seen as an adjective modifier to its noun, “Mohammed Siad Barre,” forming a phrase like “warlord Barre.” In this context, the framing component emerges from a meaningful semantic pair, a modifying word and its head noun, which together construct the framing meaning.

Crucially, this pair of words is bound by a meaningful semantic relation. For example, in the dependency parsing output of natural language processing, the noun “Barre” and its modifier “warlord” are linked by an adjectival modifier relation (known as “amod”). Based on this linguistic structure, this current research conceptualizes a framing component as “a pair of words connected by a meaningful semantic relation.” The modifying word may belong to various parts of speech, such as adjectives (e.g., young shooter), verbs (e.g., shooter kills), or even participles and modal verbs (e.g., shooter accused of [killing]).

In qualitative textual and quantitative content analyses that rely on manual labor, scholars might code the keyword “warlord,” keeping other parts (e.g., noun and semantic relation) in mind, and consider its semantic context during interpretation to explore meaningful insights. However, for computational analysis, capturing such semantic structures explicitly becomes essential for scaling framing analysis to large datasets. 

To this end, this current research operationalizes a framing component as a pair of words connected by a meaningful semantic relation, specifically identified using dependency parsing techniques. For instance, adjective-noun (amod) or verb-subject (nsubj) relationships are used to detect modifier-entity structures, such as “teenage gunman” or “shooter kills.” These semantic relations are computationally extracted from the dependency tree of each sentence. By identifying the framing components in the semantic relations-based structure, this approach allows for systematic extraction of entity-modifier pairs in large datasets, ensuring both consistency and scalability. This operationalization is particularly well-suited for analyzing entity-centric frames, as it captures how individuals, organizations, or groups are framed through specific modifying words in large datasets.

\subsection{Framing Analysis with Computational Approaches}

Traditionally, researchers utilize qualitative and quantitative methods to analyze frames, relying on manual labor and small amounts of data ~\cite{d2018doing, reese2001framing}. To tackle the challenge of analyzing frames in large-scale datasets, scholars have begun using computational approaches—both supervised and unsupervised—in the last two decades ~\citep[e.g.,][]{card2015media, liu2019detecting, walter2019news, van2018communication}.  

\textbf{Supervised.} A supervised approach needs pre-labeled datasets. In this approach, a model is first trained on the labeled data and then applied to a new dataset to classify or predict each instance ~\cite{kotsiantis2007supervised}. Under the supervised framing analysis approach, ~\cite{liu2019detecting} proposed a deep learning-based model developed with manual codes of headlines of news reports relating to gun violence. 

\textbf{Unsupervised.} An unsupervised approach does not require any pre-annotated datasets. Instead, it inductively explores all unlabeled data ~\cite{kotsiantis2007supervised}. Existing unsupervised approaches used to analyze frames include topic modeling ~\cite{dimaggio2013exploiting}, structural topic modeling ~\cite{gilardi2021policy}, hierarchical topic modeling ~\cite{nguyen2015guided}, cluster analysis ~\cite{burscher2016frames}, frequency-based models ~\cite{sanderink2020shattered}, and FrameAxis ~\cite{kwak2021frameaxis}. Compared to supervised models, unsupervised ones demand less time and can be replicated across domains. 

\textbf{\textit{Semantic relations.}} Existing unsupervised computational approaches for framing analysis are mainly based on the ideas of frequency and co-occurrences of words, resulting in the identification of discussion topics or themes, instead of frames ~\cite{ali2022survey}. Such topics do not provide a coherent framing interpretation. As per the framing conceptualization ~\cite{entman1993framing, reese2001framing}, semantic relations among words are a key to going deeper into frames, compared to the current bag-of-words-based practices, such as topic modeling. This limitation calls for exploring an unsupervised technique to capture semantic relations among words for better identifying frames. This article intends to fill the gap by focusing on unsupervised methods of framing analysis.

Although a few studies attempted to address the task with semantic relations, their approaches are not sufficiently comprehensive or supervised from the data analysis perspective. For example, ~\cite{sturdza2018automated} describes an approach of operationalizing frames using a rule-based system with a software named TurboParser. However, the author did not execute it using a dataset, leaving its usefulness unclear. A recent study by ~\citet{ziems2021protect} proposes an NLP framework to understand the frames of an entity or issue (e.g., victims in police violence) with relevant attributes (e.g., age, gender, race). However, they pre-determined the attributes and then string-matched relevant tokens as a way of framing particular entities, which is also considered supervised. 

Another study by ~\citet{van2013automatically} presents a computational framing analysis method based on semantic relations. Their approach is also a kind of supervised task, as it first determines and labels particular frames and then identifies occurrences of each pre-determined frame in the dataset. Framing analysis scholars in recent studies ~\citep[e.g.,][]{nicholls2021computational, ali2022survey} call for exploring semantic relations for improved framing nuances. 

Therefore, this research seeks to fill the gap by offering and advancing a semantic relations-based unsupervised approach for framing analysis through two studies—qualitative textual analysis and computational analysis. Both studies examine a sample of 100 news reports published by four major U.S. news media outlets on the 2022 Texas school mass shooting.

\subsection{Gun Violence and Framing Analysis}

Gun violence is a widely studied area in the U.S., as the mass shooting problem has been on the rise for years ~\cite{elbawab2022}. The body of gun violence research involves various other issues, such as mental illness ~\cite{mcginty2014news}, frames ~\cite{morin2016framing}, and public health issues ~\cite{mckeever2022gun}. Analyzing a sample of news articles on serious mental illness and gun violence, ~\citet{mcginty2014news} found that "dangerous people" with serious mental illness were more likely to be mentioned as a cause of gun violence than “dangerous weapons.” A recent study by ~\citet{mckeever2022gun} conducted an online survey (N=510) and found gun control and gun rights as the two salience frames. They also revealed that people held individuals responsible for gun violence and identified background checks as the most salient solution. 

\subsection{Attribution Theory}

The root of frames is drawn from the assumptions outlined in attribution theory (AT) ~\cite{heider2013psychology, kelley1973processes, pan1993framing}. So, this research analyzes and explains frames through the lens of AT. Originally developed within social psychology, the theory primarily describes how people explain and perceive the causes of an individual’s behavior ~\cite{heider2013psychology, mcleod2010}. While defining the theory, ~\citet{kelley1973processes} says: 

\begin{quote}
Attribution theory is a theory about how people make causal explanations, about how they answer questions beginning with "why?" It deals with the information they use in making causal inferences, and with what they do with this information to answer causal questions. (p. 107).
\end{quote}

As naïve psychologists, people tend to make two broad types of causal attributions: a) dispositional attributions and b) situational attributions ~\cite{heider2013psychology, kelley1973processes}. Dispositional attributions point to an individual’s internal factors as being responsible for an incident. For example, in a car crash, labeling people’s reckless driving behavior as a cause could be a dispositional attribution. Situational attributions refer to factors that exist outside an individual and are prevalent in specific situations. In the same example, attributing the snowy road as a cause could be considered a situational factor. Two prominent frameworks provide potential factors and insights that shape people’s perceptions of dispositional and situational attributions. These are the covariation model ~\cite{kelley1973processes} and the correspondent inference ~\cite{jones1965acts}. 

\subsection{Covariation model}

The covariation model of ~\citet{kelley1973processes} identified three potential factors leading to causal perceptions. These are consensus, distinctiveness, and consistency. 1) Consensus is related to a person or entity that explains how many individuals behave in the same way. High consensus indicates a higher level of situational attribution. 2) Distinctiveness is related to the situations that explain how an individual behaves in other similar situations. High distinctiveness indicates a higher level of situational attribution. 3) Consistency is related to time, which explains how frequently an individual’s behavior occurs. High consistency indicates a higher level of dispositional attributions ~\cite{kelley1973processes}.

\subsection{Correspondent inference}
~\citet{jones1965acts} offered three key factors in inferring causal attributions. 1) People’s degree of choice: A freely chosen behavior is considered to infer an individual’s dispositional attributions compared to forced behavior. 2) Social desirability of behavior: An individual’s behavior that is low in social desirability or social expectedness is more likely to make dispositional attributions compared to high social desirability. 3) Intended consequence of behavior: People infer an individual’s behavior as dispositional, especially when the behavior’s intended consequence is negative and harmful to people. 

\subsection{Case: 2022 Uvalde School Shooting} 

This study analyzes media coverage of a mass shooting that occurred on May 24, 2022, in Uvalde, Texas. An 18-year-old former student named Salvador Ramos entered Robb Elementary School with an AR-15-style rifle and opened fire ~\cite{Sandoval2023nyt, peckgoodman2022nyt}. The shooting resulted in the deaths of 19 students and two teachers and the injuries of 17 others ~\cite{jacoboelbawab2022, peckgoodman2022nyt, massshootingnd}. The Uvalde school shooting is one of the deadliest shootings in the United States in terms of the number of casualties ~\citet{massshootingnd}. 

The mass shooting incident received extensive coverage in local, national, and international news media ~\cite{kellner2025uvalde}, sparking outrage and reigniting long-standing debates over gun control and school safety and calls for action ~\cite{livingston2022}. News media coverage of the Uvalde shooting evolved over time ~\cite{kellner2025uvalde}. Soon after the incident, the then-President of the United States, Joe Biden, visited Texas to console the victims and pledged to act ~\cite{livingston2022, bidenremarks2022}. Within a month of the Uvalde school mass shooting that occurred 10 days after another shooting in Buffalo, New York, a gun safety legislation was passed by the Senate and Congress and then signed by the President on June 25, 2022. The gun safety law is reported as the first of its kind in the previous 30 years ~\cite{clydemirandal2022}. As the deadliest mass shooting in recent years and drawing widespread media coverage, the Uvalde elementary school shooting has been purposively selected for this study.

\section{Study 1: Qualitative Textual Analysis}

This study focuses on an in-depth examination of the usage patterns of specific words, such as adjectives and adverbs, and their semantic relations in constructing frames. Typically, computational tools and traditional research methods, such as qualitative and quantitative methods, are broadly pursued as separate lines of inquiry into frames. However, this study seeks to bridge this divide by utilizing the insights of inductive qualitative research to inform computational approaches in framing analysis. 

For this analysis, we purposively selected the 2022 mass shooting as a case that took place at Robb Elementary School. Specifically, we looked at how news media outlets in the right-leaning (a.k.a. WSJ and Fox News) and left-leaning categories (a.k.a. NYT and CNN) use selected modifying words (e.g., adjectives and adverbs) structured in a semantic pattern to frame the shooter, victims, and the event. 

Therefore, the following research questions are asked for exploration:

\textbf{RQ1:} How do right-leaning and left-leaning news media outlets use words and phrases to construct frames while covering the 2022 mass shooting at Robb Elementary School in Texas? 

\textbf{RQ2:} How do the right-leaning and left-leaning news media outlets frame the shooter, victims, and the mass shooting event at the Robb Elementary school in Texas? 

\textbf{RQ3:} How do the semantic relations of words in the Texas mass shooting news reports inform the computational analysis of frames?

\section{Study 1 Method}

To answer RQs, study 1 used qualitative textual analysis, a widely used approach to analyze frames inductively ~\cite{hertog2001multiperspectival}. It fits with the study’s purpose of inductively analyzing news reports to gain an in-depth understanding of frames, word usage patterns, and their semantic relations to constructing frames ~\cite{entman1993framing}. Qualitative textual analysis is “all about language, what it represents and how we use it to make sense of our [social realities]” ~\cite[p. 203]{brennen2017qualitative}. While exploring "how texts operate to produce meaning” ~\cite[p. 53]{browne2009close}, the qualitative analysis helps “make an educated guess at some of the most likely interpretations that might be made of that text” ~\cite[p. 1]{mckee2001beginner}. 

\subsection{Data Collection}

We collected a total of 100 news reports, including 600 news headlines and paragraphs, published by four news media outlets on the 2022 Robb Elementary School shooting in Texas. Each of them includes ten news reports on the shooting that took place on May 24, 2022. Of the news outlets, \textit{The New York Times} (NYT) and \textit{Cable News Network} (CNN) are selected as the left-leaning news media, and \textit{the Wall Street Journal} (WSJ) and \textit{Fox News} as the right-leaning news media ~\cite{mbfcnd}. The news media outlets were categorized based on their bias scores provided by Media Bias/Fact Check (MBFC). The MBFC is a non-partisan American independent site that provides bias scores for media outlets ~\cite{mbfcnd, odhner2024}. 

It is important to acknowledge that although \textit{Fox News} and \textit{The Wall Street Journal} are both considered right-leaning according to MBFC scores, they differ in tone and editorial focus. As noted by Ad Fontes Media (2024), \textit{Fox News} tends to be more partisan in its opinion content, whereas \textit{The Wall Street Journal} is comparatively more centrist in its news reporting ~\cite{adfontesmedia_n.d.} (“Ad Fontes Media,” n.d.). For the purposes of this study, both outlets were categorized on the right side of the media bias spectrum.

Using purposive sampling, we used the keywords “(Texas OR Uvalde) AND (“shoot*”)” and searched articles on Factiva, a global news database, for these four news media outlets separately from May 24 to 31, 2022. After downloading all news reports identified during the period, we manually scrutinized and removed articles that are not specifically relevant to this Uvalde mass shooting and are other types of content, such as editorials. This continued until 25 news reports were selected for each new media. Collecting news reports continued until the dataset seemed saturated with relevant words and phrases coded in this study. Data saturation is considered a useful guide for sampling data in a qualitative study that deals with a relatively small amount of information ~\cite{sandelowski1995sample, brennen2017qualitative}. With this process, a sample of 100 news articles was finally selected for this analysis. Since the lead and initial paragraphs(s) generally represent the most important messages in a news story ~\cite{liu2019detecting, van1985structures}, we purposively selected the headlines and first five paragraphs of each news report, totaling 600 headlines and paragraphs, for an in-depth analysis.

\subsection{Data Analysis}

This study analyzed the news reports with word-by-word coding in three phases—open coding, axial coding, and selective coding ~\cite{saldana2021coding} using NVivo, a qualitative data analysis software. The coding process was guided by the three research questions, framing theory ~\cite{entman1993framing}, and attribution theory ~\cite{heider2013psychology, kelley1973processes}. Following the research purpose and questions, the data analysis focused on using words and their semantic relations in constructing frames ~\cite{entman1993framing} in the case of the Robb Elementary School shooting. The analysis explores whether and how the news reports used various words and phrases to promote particular interpretations or evaluations relating to the shooter, victims, and the event. 

During the open coding phase, we specifically looked at the use of words and phrases that promoted or highlighted four aspects: a) the shooter, Salvador Ramos; b) victims, such as school children and teachers; c) the shooting incident, in certain ways. Each type of word and phrase was coded into a separate code. For instance, the words “kills,” “killing,” and “killed” were coded into a single code. In the axial coding phase, where related codes are grouped into broader categories, we organized initial codes into similar categories based on shared framing purposes. Finally, during the selective coding, where central theme(s) are refined, a few broad themes emerged with adequate exemplars (Table \ref{qual_framing_components_nytcnnwsjfox}. To ensure validity, we used two strategies: data triangulation (drawing from multiple data sources) and disconfirming evidence (intentionally seeking and considering both supporting and opposing evidence from data) ~\cite{creswell201630}.

\section{Study 1 Findings}

RQ1 and RQ2: The analysis identifies distinct sets of specific words and phrases in left-leaning news outlets, such as NYT and CNN, and right-leaning outlets, such as WSJ and Fox News (see Table \ref{qual_framing_components_nytcnnwsjfox}, that frame the shooter, victims, and the shooting event differently. 

\subsection{Shooter}

\textbf{“Accused” killer.} The analysis shows that both right-leaning and left-leaning media outlets use some common verbs (e.g., kills, left dead, opened fire, and shot) while attributing the shooter’s act. Importantly, right-leaning media outlets use weaker verbs and modifiers (e.g., “is accused of shooting,” “claimed lives,” and “allegedly committed by”), which casts doubt on Salvador Ramos' crime and weakens the gravity of killing people. In contrast, left-leaning media use stronger verbs and modifiers (e.g., “shot and killed,” “burst in and killed,” and “horrifically”), which presents the incident with a higher gravity of the mass shooting act. 

Differences in identifying the shooter also exist between the two groups of media outlets. In contrast to the left-leaning media outlets, the right-leaning ones use weaker terms like “alleged gunman” and “alleged shooter.” For example, a news report published by Fox News on May 27 said, “Salvador Ramos, the alleged gunman accused of shooting his grandmother and then targeting dozens of victims….” This seems to have cast doubts regarding Salvador Ramos’ act of killing people, at least to some extent.

\textbf{Mental instabilities.} In terms of attributing the responsibility or blame to Salvador for the mass shooting, both groups of news media outlets showed their own bias. In contrast to left-leaning news media outlets, the right-leaning ones selected and highlighted Salvador’s mental and family-related instabilities, drawing attention to the social factors while presenting Salvador as responsible for the shooting. For example, the WSJ reported, “Salvador Ramos… came from a broken family and unsettled classmates and co-workers with sometimes aggressive behavior and disturbing social-media posts.”  

\subsection{Shooting Incident}

\textbf{Low vs. high severity.} The shooting incident has been found to have been presented differently in terms of its severity between the right-leaning and left-leaning news media outlets. The former has employed specific words and phrases to portray the incident as a less severe one than the latter. The event modifier columns of Table \ref{qual_framing_components_nytcnnwsjfox} demonstrate that while some words and phrases are common to both groups of news outlets, others are used to attribute the shooting incident differently. For example, the NYT used the word “slaughter” to describe the incident, while Fox News and the WSJ did not. The left-leaning media also used “terrorist attack,” which was not used by their right-leaning counterparts. In contrast, right-leaning media outlets identified the shooting as a “senseless crime.” The use of words such as “terrorist attack” and “slaughter” might trigger nodes in the human brain related to other deadly terrorist incidents, portraying the shooting as a more severe act ~\cite{collins1975spreading}. On the other hand, the use of the phrase “senseless crime” suggests a typical type of crime. Therefore, differences in the use of words have contributed to defining the mass shooting as a problem in terms of its severity between the right-leaning and left-leaning news media outlets.

\section{Victims}

\textbf{Teenager vs. older.} Divergent portrayals of the victims in the news media outlets were observed, with both left-leaning and right-leaning news media using distinct words and phrases, although some terms were commonly employed. The left-leaning media outlets specifically employed phrases such as “school children,” “elementary school children,” and “kids,” whereas the right-leaning outlets used “children” and “students.” For instance, the NYT reported on “the killing of at least 19 elementary school children in second, third, and fourth grades.” Although subtle, this contrast indicated the left-leaning outlets’ emphasis on the word “elementary,” framing the shooting incident as an attack on young children of this age group. The word “elementary” distinguishes the age range of 5-10 years from “children” and “students.” The word “elementary” emphasizes the victims’ age range more clearly, whereas terms like “students” and “children” are more general and can apply across various age groups, including older youth. By promoting the ages of the victims in distinct ways, both the left-leaning and right-leaning news media outlets presented the severity of the shooting incident and drew attention to the shooting problem differently.

RQ3: Answers to RQ1 and RQ2 highlight the use of words in constructing relevant frames. The RQ3 serves the main purpose of Study 1, which is to investigate the semantic patterns or semantic relations of those words in creating frames. As a way of answering it, this analysis provides various groups of words and phrases centering on the shooter, victims, and the event, illustrating semantic relations among the words (see Table \ref{qual_framing_components_nytcnnwsjfox}.

\section{Semantic Relations}

The above results and Table \ref{qual_framing_components_nytcnnwsjfox} provided in the analysis present two crucial aspects that contribute to the understanding of frames. Firstly, the words used to construct frames are crucial in identifying the framing components utilized by the news media outlets. Secondly, the semantic relations among the words are crucial in establishing the frames’ meaning. Semantic relations indicate how the words are interrelated and which entity the words are attributed to. Reading through only the words might provide some insights into relevant framing components, but the insights are not fully meaningful without the words’ semantic relations. When the words are read with their semantic relations, it renders particular meanings to construct frames. For instance, in the excerpt “An 18-year-old gunman on Tuesday fatally shot 19 children and two adults” from an NYT article, the semantic relation between the phrase “18-year-old” and the “gunman” (Salvador) highlights that the modifier refers to the gunman and not the children. Without considering the semantic relations, it appeared challenging to comprehend the relevant meanings of the words and subsequently construct frames.

\section{Study 1 Discussion}

\subsection{Highlight and Hide}

As the findings indicate, both left-leaning and right-leaning news media highlighted some common and different words regarding the Texas mass shooting, conforming to the framing strategy of highlighting and hiding certain aspects of the event ~\cite{d2018doing, entman1993framing, greussing2017shifting}. As a frame functions to purvey various judgments about reality ~\cite{entman1993framing, d2018doing}, the frames constructed by left-leaning and right-leaning news outlets may shape how people perceive and understand the causes of the mass shooting and influence their attitudes toward it. 

\subsection{Attribution of Responsibility}

Left-leaning news media outlets attributed more responsibility to Salvador for the mass shooting compared to right-leaning ones. As the attribution theory ~\cite{kelley1973processes} and framing theory ~\cite{entman1993framing} suggests, with highlighted salience in situational factors (e.g., broken family) in right-leaning outlets, people are more likely to attribute the shooting’s causes to situational factors. This is supported by the phrase “accused of” that right-leaning news used in presenting Salvador’s shooting. 

Right-leaning news media highlighted aspects of Salvador’s social factors, such as his broken family, which may have made his actions appear more situationally driven. As per the correspondent inference model ~\cite{jones1965acts}, such social desirability can reduce attributions of personal responsibility by shifting away from dispositional factors. Overall, the left-leaning news media reports focused on attributing the causal responsibility of the mass shooting more to Salvador, while the right-leaning news media reports went beyond Salvador’s individual responsibility to his family factors. Such causal interpretation is supported by the study of ~\cite{mcginty2014news}, which shows “dangerous people” with mental illness were more likely mentioned as a cause of gun violence than “dangerous weapons.” The study by ~\cite{mckeever2022gun} also extends evidence in support of this current study’s findings. 

\subsection{Semantic Relations for Computational Framing Analysis}

Unsupervised computational methods mostly rely on the ideas of frequencies and co-occurrences of words ~\cite{blei2012probabilistic, dimaggio2013exploiting}. These bag-of-words-based approaches are not designed to look at the semantic relations of words and end up with identifying topics, instead of frames ~\cite{ali2022survey}. The study 1 findings demonstrate that capturing semantic relations helps discern in-depth nuances in the texts through word relations and, thus, identify relevant frames. For example, in the following excerpt from a New York Times article, “An 18-year-old gunman on Tuesday fatally shot 19 children and two adults,” the semantic relations show that the phrase “18-year-old” modifies the “gunman” (aka Salvador), not children. Without knowing this semantic relation, relevant meanings of the words and, subsequently, frames do not emerge (see Table \ref{qual_framing_components_nytcnnwsjfox}).

Manual data analysis enables the researchers to identify such semantic relations and relevant frames, as presented above. Therefore, semantic relations appeared essential for having relevant meanings and frames in a text. In a computational method, being able to capture the semantic relations seems to be a one-step advancement toward better identification and analysis of frames. As identified in this study 1, the lists of words, their attributes, and semantic relations for the shooter, victims, and the event are so specific that these can be incorporated into an algorithmic model. So, this study suggests incorporating these semantic relations into computational techniques (e.g., dependency parsing) for better automatic framing analysis. As envisioned in ~\citet{nicholls2021computational} and ~\citet{ali2022survey}, this current study’s findings extended additional evidence of how semantic relations among words and phrases, instead of just bag-of-words, can better explain nuances of frames, especially in an unsupervised model.

\section{Study 2: Computational Analysis}

Study 2 of computational analysis builds on the insights and recommendations from Study 1 of qualitative textual analysis. It focuses on the potential of using dependency parsing, an NLP technique that analyzes the grammatical structure of a sentence by identifying relationships between words, such as which word modifies or depends on another. This approach aims to enhance the identification and analysis of frames computationally. Examining the same dataset of news articles from Study 1, this computational analysis explores how dependency parsing can capture the semantic relations of words and understand relevant frames. We also compare the results of the unsupervised computational model (Study 2) with those obtained through manual data analysis (Study 1) to evaluate the effectiveness of the computational approach. The findings contribute to a better understanding of the role of the semantic relations-based computational approach in analyzing frames and offer insights into the potential of using dependency parsing as a methodological approach for framing analysis.

Since it is one of the first studies to use semantic relations in analyzing frames, we offer similar research questions established in study 1, consistent with the objectives of study 2.

\textbf{RQ1:} How do right-leaning and left-leaning news media outlets use words and phrases to construct frames at the Robb Elementary school in Texas? 

\textbf{RQ2:} How do the right-leaning and left-leaning news media outlets frame the shooter, victims, and the mass shooting event at the Robb Elementary school in Texas? 

\section{Study 2 Method}
\subsection{Dataset}
To answer the research questions, we analyzed the same news report dataset as study 1. Parsed by the spaCy NLP language model, the dataset contains a total of 24604 tokens, with 4768 for CNN, 6282 for Fox, 6759 for NYT, and 6795 for WSJ. We used the same dataset to compare the frames provided by the computational approach with those of the qualitative study. 

\subsection{Analysis}
The data analysis involved the following seven steps:

\textbf{1) Coreference resolution:} As this study aims to identify modifying words centering three entities, the shooter, victims, and the event, we needed to identify and resolve the coreferences (e.g., “he” or “suspect” for the shooter) to capture all possible modifying words of both “references” and “co-references.” To accomplish this, we applied NeuralCoref, an extension of the spaCy NLP library that provides coreference resolution.

\textbf{2) Token extraction:} We then applied a dependency parser of the spaCy language model that parsed all the news reports and generated a dependency parse tree. This tree provides the syntactic structure of a sentence that includes nodes, such as heads (e.g., gunman) and children (e.g., suspect), representing words, and edges representing the semantic relationships between those heads and children. Each edge is labeled with a specific dependency relation, such as “amod” (adjective modifier).

\textbf{3) Determining keywords:} To capture all possible words that refer and co-refer to each of the three entities, we determined relevant keywords for each entity (e.g., Salvador, gunman, shooter). These keywords were determined based on study 1 insights and then refined through manual checking of some tokens in the output produced in step 2 (see Table \ref{keywords_rels_shooter_victims_events} for details). 

\textbf{4) Filtering heads and children:} Based on the keywords, we filtered out all relevant “heads” and “children” of each entity, all their dependency relations, and associated news outlets. 

\textbf{5) Determining and refining dependency relations:} This step determines and refines useful dependency relations based on this study’s purpose. We removed some dependency relations (e.g., cc, punc) that were not useful in making meanings in relation to the RQs, by manual checking of the relations grouped in the output produced in step 4 (see Table \ref{keywords_rels_shooter_victims_events} for details). 

\textbf{6) Filtering “framing components”:} We consider each pair of head and child with certain dependency relation (e.g., the pair of “shooting” keyword and “deadly” child with “amod” relation) as a framing component that provides a particular attribution to an entity. This step filtered out all framing components for each entity by the news outlets. 

\textbf{7) Framing components to frames:} Until the last step, we analyzed the data computationally using spaCy and Pandas, a popular data analysis library for Python. In this step, we followed both computational and manual qualitative explorations. 7a) Computational: We computationally grouped the framing components for each entity by dependency relations. To achieve this goal, we used BERT word embedding and k-means clustering of the modifying words (also known as children). 7b) Qualitative: We inductively coded the modifying words and categorized them into groups following the research questions manually. Here, we consider a single framing component as a candidate for being included in multiple groups (Saldaña, 2016), and triangulation and disconfirming evidence were utilized to ensure the validity (Creswell, 2016). In both parts, each group is considered as a frame. With the process, a number of frames emerged with exemplars.

\section{Study 2 Findings}
This section reports the findings of the qualitative analysis in step 7, followed by the computational analysis from steps 1 to 6. The results of the computational exploration in step 7 are not reported here, as we found that the findings from manual analysis outperformed them. The clusters revealed through k-means clustering were not found to be coherent and adequately insightful for understanding the nuances of frames, as we examined the results manually. The findings of the qualitative analysis reveal that right-leaning and left-leaning news media outlets use different words to construct frames of the shooter, victims, and the mass shooting event at the Robb Elementary school in Texas differently, as presented in Tables \ref{comp_SHOOTER_frame_components_nyt_cnn_wsj_fox}, \ref{comp_VICTIM_frame_components_nyt_cnn_wsj_fox}, and \ref{comp_EVENT_frame_components_nyt_cnn_wsj_fox} respectively. 

\subsection{Shooter: “Accused” killer}
The shooter was characterized with some words that create doubt over the shooter’s killing action. Comparing the attributions used by right-leaning and left-leaning news outlets, it was found that the former used the words "alleged [shooter]" and “suspected [shooter]” more frequently than the latter. Furthermore, a right-leaning news outlet referred to the shooter as “accused [of shooting]”, which was not used by the left-leaning outlets. These attribution differences suggest that the two media outlets utilized different priorities in framing the shooter. 

\subsection{Shooter: Diversity of attributes}
As depicted in Table \ref{comp_SHOOTER_frame_components_nyt_cnn_wsj_fox}, right-leaning news outlets used a greater variety of attributes to highlight various aspects of the shooter than left-leaning outlets. For example, right-leaning outlets used words such as “unhappy,” “deceased,” “civilized,” and “active” to describe the shooter, which left-leaning outlets did not use. However, these words appear scattered and do not seem to form a coherent argument. This may be due to the small dataset used in this study. A larger dataset in future research could reveal more modifying words and categorize them into relevant groups, providing further insights into framing strategies.

\subsection{Shooter \& Victims: Teenager vs. Older}
Right-leaning news media outlets tend to use words depicting the “shooter” as comparatively younger than left-leaning outlets. For instance, words used by the right-leaning outlets to attribute to the shooter include “teenage,” “young,” and “student,” which left-leaning outlets did not mention. Another example is that the shooter was identified as “18-year-old” 26 times in the left-leaning outlets and only 10 times in the right-leaning ones (see Table \ref{comp_SHOOTER_frame_components_nyt_cnn_wsj_fox}. In contrast, the victims were attributed with the word “young,” an adjective modifier, five times by the left-leaning outlets and zero times by the right-leaning ones. Overall, right-leaning outlets frame the shooter as younger and the victims as older, and the scenario is the opposite in left-leaning outlets (see Table \ref{comp_VICTIM_frame_components_nyt_cnn_wsj_fox}. 

\subsection{Victims: Our Kids vs. Your Kids}
There is not much difference between left-leaning and right-leaning news outlets in using personal pronouns to modify the victims (see Table \ref{comp_VICTIM_frame_components_nyt_cnn_wsj_fox}. Pronouns addressing victims are important to perceive how the news media outlets stand with them. The left-leaning outlets still used a greater variety of personal words, such as my (2), our (2), and your (2), while the right-leaning ones used two such words, her (2) and our (3). 

\subsection{Shooting Event: Low vs. High Severity}
To describe the shooting, left-leaning news media outlets tend to use more severe and emotionally charged words, such as deadliest (6), deadly (6), horrific (1), horrifying (1), heinous (1), tragic (1), and fatally [shot] (1), which frames the issue as a more significant problem. Such words used by right-leaning outlets include deadly (4), deadliest (3), awful (4), horrific (3), senseless (2), and devastated (1). This shows the right-leaning outlets use less intense words like “senseless” and “awful,” which suggests a less severe framing of the issue (see Table \ref{comp_EVENT_frame_components_nyt_cnn_wsj_fox}. Overall, the mass shooting framing is constructed by the language deployed by news media outlets, and the severity of the framing can differ based on the political leaning of the outlet, which is aligned with framing aspects suggested by ~\cite{entman1993framing}. 

\section{Study 2 Discussion}
Study 2 investigates how news outlets frame the shooter, victims, and the Texas school shooting, applying a new computational approach based on semantic relations. 

\subsection{Attribution of Responsibility}
Framing the shooter as “young” or “older” can have significant implications for how people perceive the shooting and the level of responsibility attributed to the shooter ~\cite{entman1993framing}. The use of the “young” attribute by right-leaning outlets could soften the shooter’s image and create a more sympathetic portrayal, thereby reducing the level of responsibility attributed to him ~\cite{jones1965acts}. On the other hand, the left-leaning outlets’ focus on the victims’ youth could create a greater sense of tragedy and urgency and, therefore, a higher level of responsibility attributed to the shooter ~\cite{decety2012contribution}. As per attribution theory, people tend to attribute a person's behavior to internal or external factors based on internal and external factors ~\cite{heider2013psychology, kelley1973processes}. In this case, the framing of the shooter and victims differently by the news outlets might shape how people attribute responsibility for the shooting. The framing differences among news media outlets might have been shaped more by established media routines and practices ~\cite{reese2018media} than by the specifics of this particular mass shooting event.

\subsection{Taking Actions}
The news media outlets' different approaches to highlighting selected “severe” words might have significant implications for how the public perceives the incident and “choose[s] to act upon” the problem ~\cite[p. 54]{entman1993framing}. The left-leaning news outlets' higher salience on words like “deadly” and “deadliest” might activate the "amygdala” node in people’s brains, potentially leading them to take actions like protest and advocacy ~\cite{phelps2006emotion, barry2013after}. At the same time, highlighting more on the words “accused” and “alleged” regarding the shooter’s act, the right-leaning news outlets, compared to left-leaning ones, offered doubt in people’s perception regarding Ramos’s mass shooting. Such higher salience on these words in right-leaning outlets seems to have weakened people’s perception of the shooter’s dispositional factors in committing the offense ~\cite{kelley1973processes}. 

\subsection{Highlight and Hide}
Taking some meanings or words over others as discussed above conforms the framing technique of highlight and hide, as proposed by ~\citet{entman1993framing} and ~\citet{fairhurst1996art}. Both right-leaning and left-leaning news outlets utilized distinct ways of framing the shooter, victims, and the event despite some common depictions between the groups. Overall, the left-leaning outlets attempt to elicit people’s sympathy for “victims” while right-leaning ones sympathize with the shooter, as evidenced above. 

\section{Integrated Discussion of Both Studies}

This research's primary objective is to introduce and explore a new approach to computational framing analysis. Our initial qualitative inquiry in study 1 revealed in-depth insights into the role of semantic relations in frame construction and suggested that dependency parsing, a computational method, could potentially serve as a practical unsupervised approach to frame analysis. Based on study 1's findings and recommendations, study 2 applied dependency parsing to the same dataset as an approach to computationally analyzing frames. 
A comparison of the findings of both studies demonstrates the potential of this proposed semantic relations-based approach to automate the identification and analysis of frames in large datasets. As the study 2 discussion suggests, its findings on framing the shooter, victims, and the event are well interpretable with relevant theoretical frameworks, and the interpretations are mostly aligned with those of study 1 and prior gun violence research. With that, this article proposes a novel computational framing analysis approach based on dependency parsing named \textit{“Semantic Relations-based Unsupervised Framing Analysis”} (SUFA). 

\section{Semantic Relations-based Unsupervised Framing Analysis (SUFA)}

\subsection{Novelty of SUFA}
The SUFA is novel in analyzing frames in several ways. First, it is based on semantic relations that extend beyond the bag-of-words approach utilized by most existing unsupervised computational framing analysis methods, such as topic modeling. Second, as discussed above, a few studies have employed semantic relations in framing analysis ~\citep[e.g.,][]{van2013automatically, ziems2021protect}. However, they did not present it as an unsupervised method. In contrast to these studies, our approach demonstrates its distinction as an unsupervised method. Researchers do not need to define a frame in advance to explore frames within a dataset. Third, our approach provides flexibility in using qualitative manual coding or computational tools like word embedding and k-means clustering in step 7 of its data analysis process. In this sense, it is a mixed-method approach that prior studies did not include. 

\subsection{Data Analysis in SUFA.}
The procedure for analyzing data in SUFA is outlined in seven steps in the study 2's method section. To effectively apply SUFA, we recommend following these steps along with a few additional considerations. If news media outlets have specific identities, such as left or right-leaning, we suggest labeling these identities in the data. Steps 3 and 5 require human intelligence. For example, step 3 involves providing keywords for each entity, which can be informed by domain knowledge, researchers’ little manual data exploration, or with the assistance of WordNet ~\cite{miller1995wordnet}. In step 5, researchers may need to manually review the output to identify useful semantic relations for analysis. For step 7, either qualitative or computational analyses can be used, depending on the research goals and the number of modifying words derived from the dataset. The computational analysis, such as word embedding and k-means clustering, may generally be more appropriate as SUFA is meant to analyze large datasets. However, if the size of modifying words is small enough to manage manually, a qualitative analysis might be more suitable for step 7, as research suggests that human intelligence often outperforms machines in tasks that require contextual interpretation and subjective judgment ~\cite{lazer2009computational}.

\subsection{Strengths}
The SUFA is an unsupervised approach that does not require any prior labeling or defining of data frames. Instead, it uses an inductive approach to explore and group attributions together to reveal framing components or frames. One advantage of SUFA is that it allows for the flexibility of utilizing both human intelligence and computational techniques to emerge frames and their interpretations, particularly in cases where the size of modifying words is small enough to be managed manually. Moreover, SUFA can inductively analyze frames in large datasets in an unsupervised manner.

\subsection{Weaknesses}
The approach requires manual input in determining relevant keywords (step 3) and semantic relations (step 5), which can be time-consuming and subjective. It is limited to analyzing frames centered around entities, such as an individual (e.g., shooter), a group of people or community (e.g., victims), and an incident or phenomenon (e.g., a shooting event). Since this study focuses on exploring emphasis frames through a semantic relations-based approach, it is better suited for analyzing emphasis frames ~\cite{D'Angelo2017} rather than equivalency frames ~\cite{kahneman1984choices}. 

Additionally, like other computational framing analysis approaches, this study only considers words and phrases, while other framing components like metaphor, placement, and visual elements are not analyzed. During coreference resolution (step 1), some words useful as framing components could be replaced with co-references (e.g., replacing the word “gunman” with “Salvador”), which might lead to the loss of some words with important nuances.

\section{Conclusion}
This research introduces a new computational approach called \textit{Semantic Relations-based Unsupervised Framing Analysis} (SUFA), which utilizes semantic relations to analyze news frames. While the method has some limitations, such as the need for manual input and its focus on emphasis frames, it provides a useful tool for exploring framing components in news media coverage. The mixed-method approach of SUFA offers researchers the flexibility to use entirely computational tools or couple it with qualitative manual coding, where applicable, for data analysis. Overall, SUFA is a valuable addition to the field of computational framing analysis, enabling more comprehensive and nuanced analysis of news media frames.

\section{Limitations and Future Research}
The SUFA was developed and tested on a single dataset of news reports on gun violence in Study 1 and Study 2. However, the approach can be applied to other domains with the provision of relevant keywords and relations. Further research can be conducted to explore the applicability of this method in other domains and to improve its performance. Currently, SUFA only considers words when analyzing frames. However, the computational framing analysis needs to include other framing components such as metaphor, visual content, placement, differences between headline and body texts, and exemplars. Such advancements will provide a more comprehensive understanding of framing effects in news media.

\begin{sloppypar}
\bibliographystyle{ACM-Reference-Format}
\end{sloppypar}

\bibliography{6_anthology}

% \onecolumn
% \appendix

% \clearpage
% \appendix

% \section{Appendix}

% % \documentclass{article}
% \usepackage{tabularx}
% \usepackage[margin=1in]{geometry}
% \usepackage{array}
% -------------------------------------------
% \clearpage
% \input{7_appendix_aej}

\onecolumn

\appendix

\section{Appendix}
\label{sec:appendix}
% ---------------------------

% Table 1-----------------------------------------------------------------------
% \clearpage

% \usepackage{multirow}
% \usepackage{longtable}
% Note: It may be necessary to compile the document several times to get a multi-page table to line up properly

% \begin{longtable}{|l|l|l|}
\begin{longtable}{|l|l|>{\raggedright\arraybackslash}p{12.5cm}|}
\caption{Words, Phrases, and Their Relations in News Reports of Two Media Groups: NYT and CNN vs. WSJ and FOX.}
\label{tab:my-table}\\
\hline
\multirow{2}{*}{\textbf{\begin{tabular}[c]{@{}l@{}}Salvador \\ Modifiers\end{tabular}}} &
  \textit{NYT, CNN} &
  \begin{tabular}[c]{@{}l@{}}18-year-old, 18-year-old man, Armored, Gunman, He, Shooter, Suspect\end{tabular} \\ \cline{2-3} 
 &
  \textit{WSJ, FOX} &
  \begin{tabular}[c]{@{}l@{}}18-year-old, 18-year-old man, A resident of Uvalde, Active shooter, Alleged gunman, Alleged \\ shooter, Alone, Former student at Uvalde High school, Gunman, He, Mass shooter, Now-deceased, \\ now-deceased suspect, Shooter, Suspect, Suspected lone gunman, Suspected shooter, Suspected \\Uvalde school shooter, Texas school shooting suspect\end{tabular} \\ \hline
\endfirsthead
\endhead
\multirow{2}{*}{\textbf{\begin{tabular}[c]{@{}l@{}}Salvador \\ History\end{tabular}}} &
  \textit{NYT, CNN} &
   \\ \cline{2-3} 
 &
  \textit{WSJ, FOX} &
  \begin{tabular}[c]{@{}l@{}}Broken family, Hostile, Unsettled classmates, Violent, was 'flashing red'\end{tabular} \\ \hline
\multirow{2}{*}{\textbf{Gun}} &
  \textit{NYT, CNN} &
  \begin{tabular}[c]{@{}l@{}}A long rifle, Assault rifle, Semiautomatic rifle, Semiautomatic weapons, Semitauonic firearms, \\Tactical vest, With a rifle\end{tabular} \\ \cline{2-3} 
 &
  \textit{WSJ, FOX} &
  AR-platform rifle, Handgun, Legally purchased, Two rifles \\ \hline
\multirow{2}{*}{\textbf{Verb}} &
  \textit{NYT, CNN} &
  \begin{tabular}[c]{@{}l@{}}Burst in and killed, Came in an opened fire, in {[}in custody{]}, Kills, Left, Left dead, Left killing,\\ Opened fire, Shoots, Shot and killed, Shot dead, Stormed into\end{tabular} \\ \cline{2-3} 
 &
  \textit{WSJ, FOX} &
  \begin{tabular}[c]{@{}l@{}}Accused of shooting, Allegedly committed by, Broke into the school, Claimed the lives …, \\Entering {[}the school{]}, Shot, Gunned down, Is accused of, Kills, Left, Left dead, Left killing, \\Opened fire, Walking into {[}school{]}\end{tabular} \\ \hline
\multirow{2}{*}{\textbf{Verb modifier}} &
  \textit{NYT, CNN} &
  After … {[}another event{]}, Fatally, Horrifically, Incomprehensibly \\ \cline{2-3} 
 &
  \textit{WSJ, FOX} &
  Fatally \\ \hline
\multirow{2}{*}{\textbf{Victim}} &
  \textit{NYT, CNN} &
  \begin{tabular}[c]{@{}l@{}}18, 19, Adults, Age between 6 and 7 years old, At least, Children,  Elementary school children, \\Elementary school students, Kids,  One, School children, Students, Teachers, Two, Victims\end{tabular} \\ \cline{2-3} 
 &
  \textit{WSJ, FOX} &
  \begin{tabular}[c]{@{}l@{}}14, 19, Adults, At least, Children, Children, One, Students, Teacher(s), Two, Victims, Xavier Lopez\end{tabular} \\ \hline
\multirow{2}{*}{\textbf{Event modifier}} &
  \textit{NYT, CNN} &
  \begin{tabular}[c]{@{}l@{}}30th K-12 shooting, 6-year-old son, Aftermath, Attack, Deadliest mass shooting, Deadly shooting, \\Devastating, Elementary school shooting, Horrific mass murder, Mass school shooting, \\ Mass shooting, Massacre, Nation reeling, School massacre, School shooting, Second  deadliest, \\ Shakes a nation, Slaughter, Slayings, Stealing their lives,  Terrorism, Terrorist attack, Tragedy, \\ Tragic, Violent, Worst school shooting\end{tabular} \\ \cline{2-3} 
 &
  \textit{WSJ, FOX} &
  \begin{tabular}[c]{@{}l@{}}Aftermath, Attack, Deadliest, Deadliest shooting, Deadly, Deadly {[}shooting{]}, Devastated the \\town, elementary school shooting, Horrific shooting, Horrific tragedy, Later discovered to be the \\ shooting, Local elementary-school shooting, Mass casualty incident, Mass murder, Mass shooting, \\Massacre, murders, School shooting, Senseless crime, Shocked the country, Shooting, Texas \\elementary, school shooting, Texas mass shooting, Texas school shooting, third most deadly, \\tragedy\end{tabular} 
\label{qual_framing_components_nytcnnwsjfox}\\ \hline
\end{longtable}

% Table 2 -----------------------------------------------------------------------

% \onecolumn

% \usepackage{longtable}
% Note: It may be necessary to compile the document several times to get a multi-page table to line up properly
\begin{longtable}{|l|l|l|}
% \begin{longtable}{|l|l|>{\raggedright\arraybackslash}p{13.5cm}|}
\caption{Keywords and Dependency Relations Used for the Shooter, Victims, and the Event}
\label{tab:my-table}\\
\hline
\textbf{} & \textbf{Keywords} & \textbf{Relations} \\ \hline
\endfirsthead
\endhead
Shooter &
  \begin{tabular}[c]{@{}l@{}}‘gunman’, ‘gunmen’, ‘man’, ‘Salvador’, ‘Ramos’, \\‘shooter’,   ‘shooters’, and ‘suspect’.\end{tabular} &
  \begin{tabular}[c]{@{}l@{}}'acl', "amod", 'appos', \\ "compound",   "relc", 'nsubj','dobj', and 'nsubjpass'..\end{tabular} \\ \hline
Victims &
  \begin{tabular}[c]{@{}l@{}}‘adult’, ‘adults’, ‘child’, ‘children’, ‘kids’,  ‘schoolchildren’, ‘student’, \\ ‘students’, ‘teacher’, ‘teachers’, ‘victim’, and   ‘victims’.\end{tabular} &
  \begin{tabular}[c]{@{}l@{}}'acl', 'compound', 'nummod', 'relcl', 'amod', 'dobj',\\   'nsubj', 'nsubjpass', and 'poss'\end{tabular} \\ \hline
Event &
  \begin{tabular}[c]{@{}l@{}}‘shooting’, ‘shootings’, ‘attack’, ‘massacre’, ‘event’,   ‘tragedy’, \\‘terrorism’, ‘slaughter’, ‘crime’, ‘slayings’, ‘murder’, and   ‘aftermath’.\end{tabular} &
  \begin{tabular}[c]{@{}l@{}}'amod', 'advmod', 'compound', 'nummod', and 'relcl'\end{tabular} 
\label{keywords_rels_shooter_victims_events}\\ \hline

\end{longtable}

% Table 3-----------------------------------------------------------------------
% \onecolumn
\clearpage

% \usepackage{longtable}
% Note: It may be necessary to compile the document several times to get a multi-page table to line up properly
% \begin{longtable}{|l|l|l|}
\begin{longtable}{|l|l|>{\raggedright\arraybackslash}p{14cm}|}
\caption{Framing Components (with Frequencies) Deployed by Each News Media Outlet to Attribute the SHOOTER, Grouped under Different Associated Semantic Relations.}
\label{tab:my-table}\\
\hline
% \textbf{Extreme left}  & \textit{} &  \\ \hline
\endfirsthead
\endhead
\textbf{Left} &
  \textit{CNN} &
  \begin{tabular}[c]{@{}l@{}}\textbf{acl:} clad (2), identified (2); \\ \textbf{amod:} active (3), old (21), deranged (1), many (1), other (1), suspected (1), alleged (1), grandmother (1); \\ \textbf{appos:} Ramos (2); \\ \textbf{compound:} Salvador (7), mass (1)\end{tabular} \\ \hline
\textbf{Left-center} &
  \textit{NYT} &
  \begin{tabular}[c]{@{}l@{}}\textbf{acl:} approaching (2), barricaded (2), driven (1); \\ \textbf{amod:} shooting (1), angry (1), armed (2), old (5); \\ \textbf{compound:} shooting (2), Salvador (2)\end{tabular} \\ \hline
% \textbf{Least-biased}  & \textit{} &  \\ \hline
\textbf{Right-center} &
  \textit{WSJ} &
  \begin{tabular}[c]{@{}l@{}}\textbf{acl:} named (1); \\ \textbf{amod:} grandmother (1), old (8), teenage (1), unhappy (1), young (1), deceased (2), civilized (2), active (4); \\ \textbf{appos:} himself (1), student (1), resident (1), old (1), man (1), birthday (1), Ramos (2); \\ \textbf{compound:} school (1), mass (4), Salvador (12)\end{tabular} \\ \hline
\textbf{Right} &
  \textit{FOX} &
  \begin{tabular}[c]{@{}l@{}}\textbf{acl:} accused (1), identified (1); \\ \textbf{amod:} active (1), alleged (2), bureaudefined (1), deceased (4), lone (2), old (1), suspected (4), upstate (1), red (1); \\ \textbf{appos:} resident (2), ones (1), gunman (1), Romas (1), Ramos (2), 18 (1); \\ \textbf{compound:} suspect (1), York (1), resident (1), mass (1), Texas (1), Salvador (14), Ramos (1), school (1)\end{tabular} 
\label{comp_SHOOTER_frame_components_nyt_cnn_wsj_fox}\\ \hline
% \textbf{Extreme right} & \textit{} &  \\ \hline

\end{longtable}

% Table 4-----------------------------------------------------------------------

% \clearpage
% \usepackage{longtable}
% Note: It may be necessary to compile the document several times to get a multi-page table to line up properly
% \begin{longtable}{|l|l|l|}
\begin{longtable}{|l|l|>{\raggedright\arraybackslash}p{14cm}|}
\caption{Framing Components (with Frequencies) Used by Each News Media Outlet to Attribute VICTIMS, Grouped under Different Associated Semantic Relations.}
\label{tab:my-table}\\
\hline
% \textbf{Extreme left} &
%   \textit{} &
%    \\ \hline
% \endfirsthead
% %
% \endhead
% %
\textbf{Left} &
  \textit{CNN} &
  \begin{tabular}[c]{@{}l@{}}\textbf{acl:} aged (1); \\ \textbf{amod:} local (1), young (2); \\ \textbf{compound:} Parents (1), parents (1), school (1); \\ \textbf{nummod: }13 (3), 14(4), 18 (1), 19(13), 20 (2), 26 (2), 535 (1), Eighteen (1), Nineteen (5), Two(2), one(3), two(17); \\ \textbf{relcl:} treated (1)\end{tabular} \\ \hline
\textbf{Left-center} &
  \textit{NYT} &
  \begin{tabular}[c]{@{}l@{}}\textbf{acl:} killed (1); \\ \textbf{amod:} dead (1), other (2), several (2), young (3); \\ \textbf{compound:} Hook (1), Uvalde (1), daughter (1), grade (1), parents (2), roll (1), school (7); \\ \textbf{nummod:} 14 (2), 18 (1), 19 (14), 20 (3), one (1), two (12); \\ \textbf{poss:} America (1), Her (1), my (2), our (2), your (2); \\ \textbf{relcl:} killed (1)\end{tabular} \\ \hline
% \textbf{Least-biased} &
%   \textit{} &
%    \\ \hline
\textbf{Right-center} &
  \textit{WSJ} &
  \begin{tabular}[c]{@{}l@{}}\textbf{acl:} celebrating (1), killed (1); \\ \textbf{amod:} former (1), other (1), small (1); \\ \textbf{compound:} Elementary (1), Robb (1), Trump (1), adult (1), mother (1); \\ \textbf{nummod:} 16 (1), 17 (1), 19 (15), 20 (1), 21 (3), four (1), two (13); \\ \textbf{poss:} her (2)\end{tabular} \\ \hline
\textbf{Right} &
  \textit{FOX} &
  \begin{tabular}[c]{@{}l@{}}\textbf{amod:} dead (1), innocent (1), little (1), ofentry (1), old (1); \\ \textbf{compound:} School (1), asa (1), center (1), school (1); \\ \textbf{nummod:} 14 (2), 18 (3), 19 (8), 4,000 (1), Two (1), eight (1), one (3), two (7); \\ \textbf{poss:} our (3); \\ \textbf{relcl:} missing (1)\end{tabular} 
\label{comp_VICTIM_frame_components_nyt_cnn_wsj_fox}\\ \hline
% \textbf{Extreme right} &
%   \textit{} &
   % \\ \hline
\end{longtable}

% Table 5-----------------------------------------------------------------------
\clearpage

% \usepackage[normalem]{ulem}
% \usepackage{longtable}
% Note: It may be necessary to compile the document several times to get a multi-page table to line up properly
% \begin{longtable}{|l|l|l|}
\begin{longtable}{|l|l|>{\raggedright\arraybackslash}p{13.5cm}|}
\caption{Framing Components (with Frequencies) Used by Each News Media Outlet to Attribute the EVENT, Grouped under Different Associated Semantic Relations.}
\label{tab:my-table}\\
\hline
% \textbf{Extreme left} &
%   \textit{} &
%    \\ \hline
% \endfirsthead
% %
% \endhead
% %
\textbf{Left} &
  \textit{CNN} &
  \begin{tabular}[c]{@{}l@{}}\textbf{advmod:} ago (1), fatally (1), least (2), \\ \textbf{amod:} 30th (2), American (2), Deadly (3), deadliest (2), deadly (2), heinous (1), horrific (1), previous (1), \\ second (2), tragic (1); \\ \textbf{compound:} Hook (2), mass (5), school (10); \\ \textbf{nummod:} 39 (2), three (1); \\ \textbf{relcl:} happened (1), left (4)\end{tabular} \\ \hline
\textbf{Left-center} &
  \textit{NYT} &
  \begin{tabular}[c]{@{}l@{}}\textbf{advmod:} ago (3), far (1); \\ \textbf{amod:} Latest (1), deadliest (6), deadly (1), horrifying (1), immediate (1), mass (7), next (1), previous (1), \\recent (1), reported (1), such (1); \\ \textbf{compound:} Buffalo (1), Newtown (1), School (2), mass (9), school (14); \\ \textbf{nummod:} 2012 (1), 215 (1), 693 (1), two (1); \\ \textbf{relcl:} killed (3), say (1), took (1)\end{tabular} \\ \hline
% \textbf{Least-biased} &
%   \textit{} &
%    \\ \hline
\textbf{Right-center} &
  \textit{WSJ} &
  \begin{tabular}[c]{@{}l@{}}\textbf{advmod:} away (2); \\ \textbf{amod:} awful (4), deadliest (1), horrific (1), latest (2), local (1), mass (2), new (2), next (1); \\ \textbf{compound:} Mass (1), mass (20), school (8); \\ \textbf{nummod:} 2011 (1), claimed (1), died (1), have (1), is (1), rises (1), targeted (1), tolerated (1)\end{tabular} \\ \hline
\textbf{Right} &
  \textit{FOX} &
  \begin{tabular}[c]{@{}l@{}}\textbf{amod:} deadliest (2), deadly (4), fourth (1), horrific (2), last (1), major (1), mass (1), recent (1), senseless (2); \\ \textbf{compound:} Parkland (1), Texas (1), Tuesday'smass (1), mass (16), preventmass (1), school (9); \\ \textbf{nummod:} 20 (1), 2018 (1); \\ \textbf{relcl:} devastated (1), had (1), happened (1), left (2)\end{tabular} 
\label{comp_EVENT_frame_components_nyt_cnn_wsj_fox}\\ \hline
% \textbf{Extreme right} &
%   \textit{} &
   % \\ \hline

\end{longtable}








\end{document}